\documentclass[conference]{IEEEtran}
\IEEEoverridecommandlockouts
\usepackage{cite}
\usepackage{amsmath,amssymb,amsfonts}
\usepackage{algorithmic}
\usepackage{graphicx}
\usepackage{textcomp}
\usepackage{xcolor}

\usepackage{amsmath,amssymb,amsfonts}
\usepackage{multirow}
\usepackage{makecell}
\usepackage{colortbl}

\definecolor{mygrey}{rgb}{.215,.215,.215}
\usepackage{mathrsfs}
\usepackage{hhline} 

\usepackage{threeparttable}  

\usepackage{booktabs}
\usepackage{tabularx}
\def\BibTeX{{\rm B\kern-.05em{\sc i\kern-.025em b}\kern-.08em
    T\kern-.1667em\lower.7ex\hbox{E}\kern-.125emX}}
\begin{document}

\title{A High-Throughput Spiking Neural Network Processor Enabling Synaptic Delay Emulation

\author{
    \IEEEauthorblockN{Faquan Chen, Qingyang Tian, Ziren Wu, Rendong Ying, Fei Wen, Peilin Liu\IEEEauthorrefmark{1}} 
    
    \IEEEauthorblockA{The School of Electronic Information and Electrical Engineering, Shanghai Jiao Tong University, Shanghai, China} 
	}
    \thanks{\hspace{-1em}\rule[0.1ex]{1\linewidth}{0.5pt} This work was supported in part by the Science and Technology Innovation (STI) 2030-Major Project under Grant 2022ZD0208700 and in part by Shanghai Municipal Science and Technology Major Project under Grant 2021SHZDZX0102. Email: fqchen1998@sjtu.edu.cn, liupeilin@sjtu.edu.cn}}
\maketitle

\begin{abstract}
    Synaptic delay has attracted significant attention in neural network dynamics for integrating and processing complex spatiotemporal information. This paper introduces a high-throughput Spiking Neural Network (SNN) processor that supports synaptic delay-based emulation for edge applications. The processor leverages a multicore pipelined architecture with parallel compute engines, capable of real-time processing of the computational load associated with synaptic delays. We develop a SoC prototype of the proposed processor on PYNQ Z2 FPGA platform and evaluate its performance using the Spiking Heidelberg Digits (SHD) benchmark for low-power keyword spotting tasks. The processor achieves 93.4\% accuracy in deployment and an average throughput of 104 samples/sec at a typical operating frequency of 125 MHz and 282 mW power consumption.
\end{abstract}
\section{Introduction}
Time serves as a critical component in Spiking Neural Networks (SNNs) that facilitate complex information processing \cite{lif}. The timing of spike emissions and synaptic delays is essential for SNN temporal modeling. Accurately simulating synaptic delays is vital for modeling biological neural systems and achieving biological plausibility and brain-like intelligence. Recent studies \cite{hammouamri} have developed methods to learn delays during training (e.g., using exponential kernels). However, implementing these delays incurs significant complexity and resource overhead, posing challenges for low-cost edge deployment. Consequently, most edge neuromorphic hardware (e.g., ReckOn \cite{reckon}, ODIN \cite{odin}) omits synaptic delay support due to cost constraints, limiting their applicability. This work develops a SNN processor capable of emulating synaptic delays based on a keyword spotting benchmark evaluation, which is typically required for always-on edge devices. 
\section{Synaptic Delay-based LIF Model}

\begin{figure}[tbp]
    \centering
    \includegraphics[width = 3.4in]{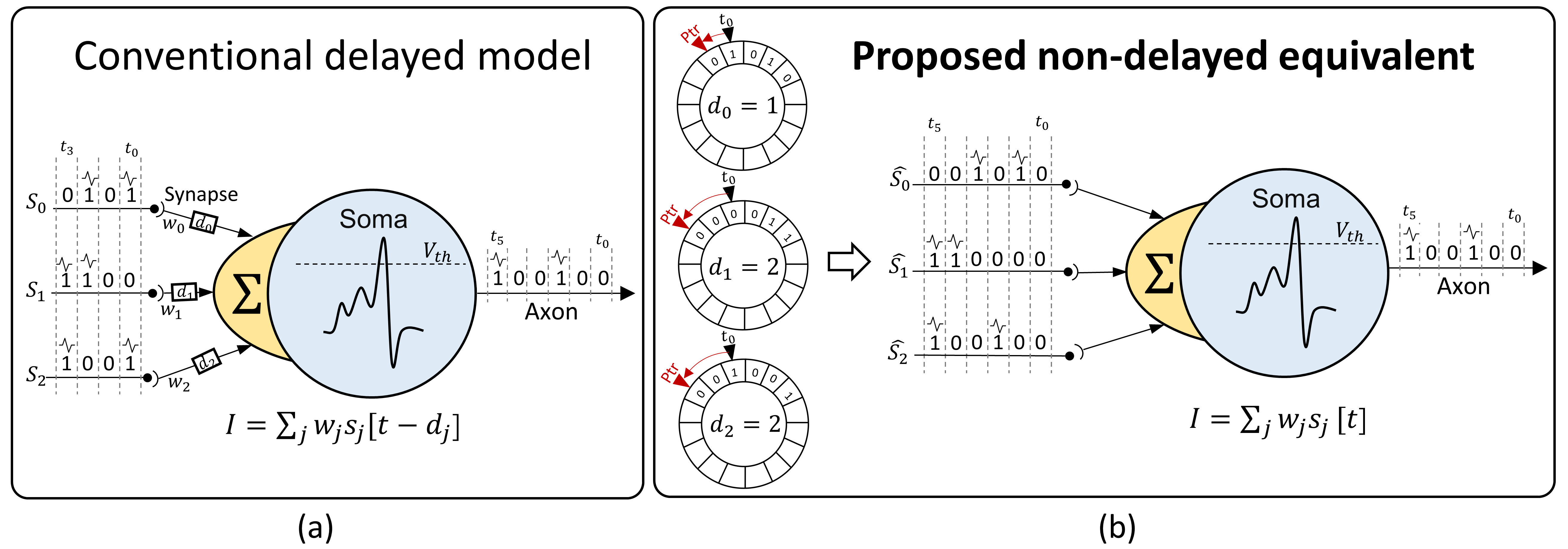}
    \caption{Comparison of synaptic delay implementation. (a) conventional delayed model. (b) proposed non-delayed equivalent.}
    \label{model}
\end{figure}

The Leaky Integrate-and-Fire (LIF) model \cite{lif} updates postsynaptic membrane potential as:
\begin{equation}
    u^{l}_i[t] = \lambda u^{l}_i[t-1] + \sum_j{w_{ij}s_j^{l-1}[t-d_{ij}]} - v_{th}s_i^{l}[t-1],
    \label{eq1}
\end{equation}
\begin{equation}
    s_i^l[t] = H(u_i^l[t]-v_{th}),
    \label{eq2}
\end{equation}
where $\lambda$ is the leakage factor, $s_j^{l-1}[t-d_{ij}]\in\{0,1\}$ is the delayed presynaptic spike, and $d_{ij}$ denotes the synaptic delay. When $u^{l}_i > v_{th}$, $s_i^l[t]$ fires via the Heaviside function $H(x)$ in Eq.\eqref{eq2}. This work converts delayed computations to non-delayed equivalents using spiking ring buffer (Fig. \ref{model}). 
\section{Hardware Architecture}
\begin{figure*}[tbp]
    \centering
    \includegraphics[width = 6.0in]{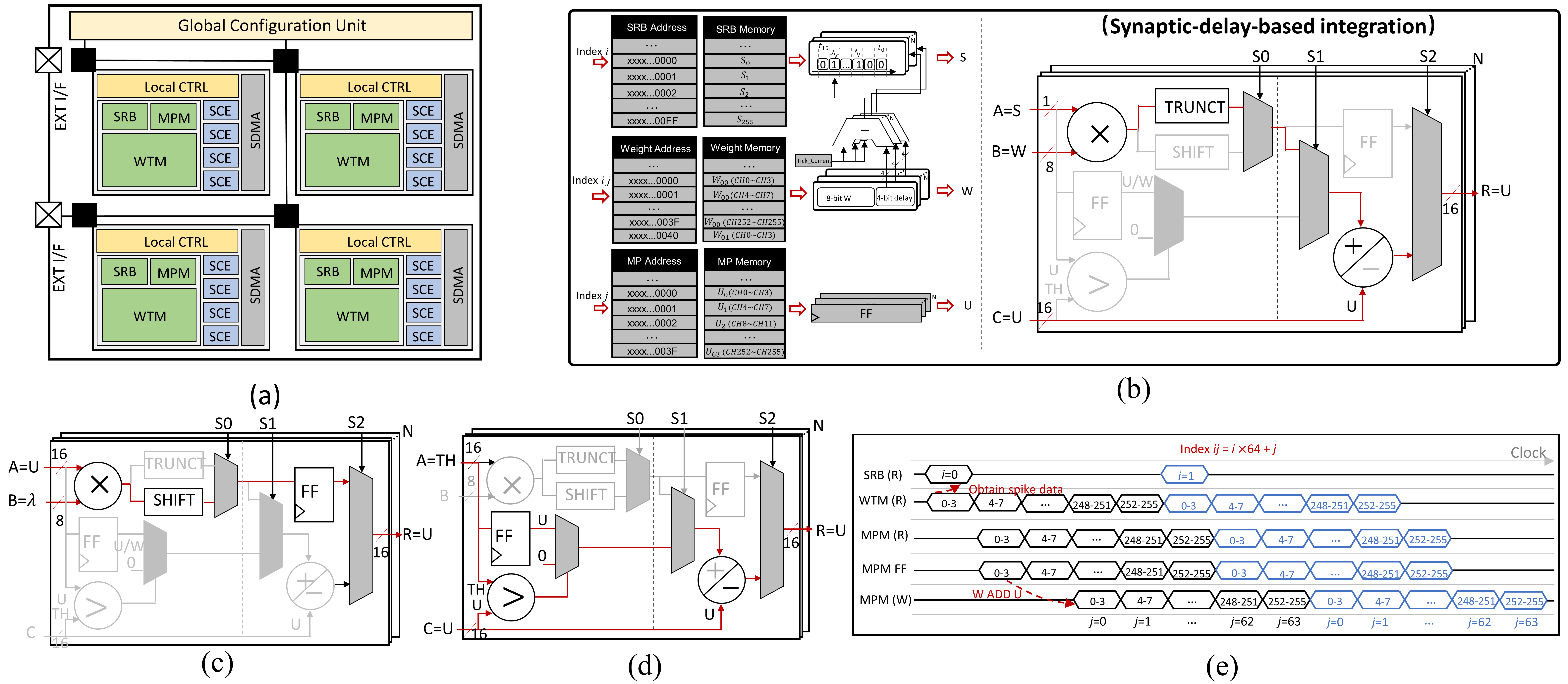}
    \caption{The overall architecture. During computation, the head pointer advances to store new presynaptic spikes at the current timestep. Synaptic weights/delays are fetched, enabling spike pointer calculation (via delay + head pointer) to access historical spikes for current-timestep processing.}
    \label{arc}
\end{figure*} 
Fig. \ref{arc} shows the overall microarchitecture of the proposed synaptic delay-based processor. The overall architecture consists of a configuration unit, external data/configuration interfaces, and four homogeneous spiking computation cores. Each computation core comprises a local controller, a memory pool, parallel Spiking Computation Engines (SCE), and a Spiking Direct Memory Access (SDMA) engine. The memory pool primarily includes a Spiking Ring Buffer (SRB), a Weight Memory (WTM) for storing weight parameters and synaptic delay, and a Membrane Potential Memory (MPM). The SRB simulates synaptic delay, with its 4-bit implementation supporting up to 15 delay units. The computation engine includes configurable data paths, fixed-point multipliers, adders, comparators, and delay units. Weights and membrane potentials are stored as 8-bit and 16-bit values, respectively. Adopting an axonal approach, computation employs dual nested loops: outer ($i$, presynaptic neurons) and inner ($j$, postsynaptic neurons). For each $w_{ij}$: (1) read SRB entry (neuron $i$'s delay window) once, (2) update connected neurons $j$ in parallel. Four postsynaptic neurons are updated in parallel, matching the engine's four channels. The computation phase consists of: membrane potential leakage, synaptic-delay-based membrane potential update, and neuron activation/reset, and corresponding computing engine configuration and timing are detailed in Fig. \ref{arc} (c), (b), (d), and (e) respectively.
\begin{table}[ht]  
    \centering  
    \caption{Comparison of Different Neuromorphic Architectures}  
    \scriptsize  
      \begin{tabular}{cccc}
        \hline
        Work & \cite{reckon} & \cite{denram} & \textbf{This Work} \\
        \hline 
        Technology & 28nm & 130nm + RRAM & \textbf{PYNQ Z2 FPGA} \\
          
        \multirow{2}{*}{Resources}  & 0.45mm\textsuperscript{2}; & \multirow{2}{*}{-} & \textbf{346kB BRAM;17DSP}\\
        & 138kB SRAM &  & \textbf{40,521LUT; 44,161FF} \\
          
        Synapses & 132k & 224k &\textbf{132k}
        \\
          
        \multirow{2}{*}{Network} & RNN& FNN+Delay & \textbf{FNN+Delay} \\
        & (256)-256-10 & (700$\times$16)-20 & \textbf{(256)-256-256-20}
        \\
          
        Training Alg. & Mod. e-prop & BPTT & \textbf{DCLS-Delays} \\
          
        \multirow{2}{*}{Accuracy} & 90.7\% & 87.5\% & \textbf{93.4\%} \\
        & @10 classes & @20 classes & \textbf{@20 classes} \\
          
        Power  & 79$\mu$W & 8.41$\mu$W & \textbf{282mW} \\
        
        Frequency & 13MHz & N/A & \textbf{125MHz} \\
         
        Latency/sample & 55.3 ms& 750 ms& \textbf{9.6 ms} \\
         
        Throughput & 18.1 (13.9$\times$)  & 1.3 (1$\times$) & \textbf{104.2 (80$\times$)} \\
        \hline
    
    \end{tabular}  
    \label{tab1}  
\end{table}  
\begin{figure}[tbp]
    \centering
    \includegraphics[width = 3.1in]{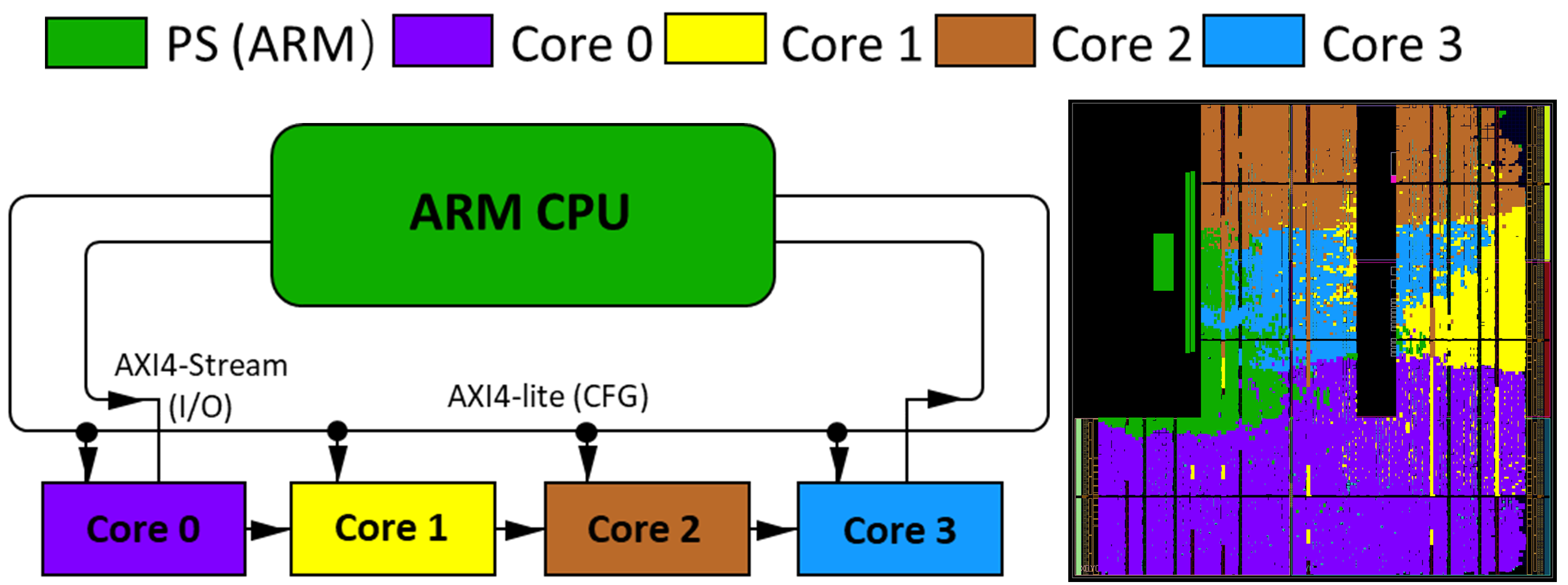}
    \caption{Placement distribution of compute cores and PS in the FPGA}
    \label{hw}
\end{figure} 
\section{Experimental Result}
The proposed processor is evaluated on the Spiking Heidelberg Digits (SHD) benchmark \cite{shd} for keyword spotting tasks, which includes approximately 10,000 spiking recordings (8,156 for training and 2,264 for testing) across 20 classes representing digits 0 to 9 in both German and English. Subsampling is performed by binning every 5 input neurons and every 10 ms. The 3-layer feedforward network is configured as (140)-256-256-20. Training follows the DCLS-Delays method \cite{hammouamri}, which is conducted over 100 time steps, with a maximum delay of 150 ms (15 units), achieving a peak accuracy of 94.7\%. After 8-bit weight symmetric quantization, an accuracy of 93.4\% is achieved for deployment.
\par The proposed processor is prototyped as a SoC on the PYNQ Z2 FPGA, which hosts a dual-core ARM Cortex-A9 CPU and 512 MB DDR3 SDRAM. We leverage the DMA configuration on the PS side of PYNQ platform to transfer spike/weight data and computation result, while the AXI-Lite interface is used for network model configuration. The SoC consumes a total of 40,521 LUTs (76.17\%), 44,161 FFs (41.50\%), 77 BRAMs (55.00\%), and 17 DSPs (7.73\%) after layout implementation (Fig. \ref{hw}). At 125 MHz, the average processing time for a single time step is 0.134 ms, much shorter than the 10 ms input time step. With 72 inference time steps per sample, the SoC achieves a high throughput of 9.6 ms/sample and 104 samples/second. The average power consumption of the processor is 282 mW at 125 MHz, with the total power consumption of the SoC at 1.71 W, and the ARM CPU consuming 1.28 W dynamic power. Table \ref{tab1} demonstrates comprehensive comparisons with \cite{reckon, denram}. Overall, the proposed SoC platform demonstrates a cost-effective method for achieving high throughput and low latency in synaptic delay-based SNNs on a reconfigurable device.
\section{Conclusion}
This work proposed an efficient SNN processor based on synaptic delay computation, aimed at enhancing the efficiency and flexibility of SNNs in handling complex temporal tasks. 

\bibliographystyle{IEEEtranS}
\bibliography{refs}

\end{document}